%% file: cvmpaper_finalcopy.tex
\documentclass[10pt,twocolumn,letterpaper]{article}

\makeatletter
\def\input@path{{body/}}
\makeatother
\usepackage{cvm}
\usepackage{times}
\usepackage{epsfig}
\usepackage{graphicx}
\usepackage{amsmath}
\usepackage{amssymb}

\usepackage{booktabs}
\usepackage{tabularx}
\usepackage{listings}
\usepackage{algorithm}
\usepackage{algorithmic}
\usepackage{caption}

\graphicspath{{.}}

\usepackage[pagebackref=true,breaklinks=true,letterpaper=true,colorlinks,bookmarks=false]{hyperref}
\usepackage[nameinlink,noabbrev]{cleveref}

\newcommand{\keywords}[1]{{\bf \emph{Keywords: #1}}}
\newcommand{\revised}[1]{#1}

\cvmfinalcopy

\def\cvmPaperID{639} 

\ifcvmfinal\fi
\begin{document}

\title{SkeletonGaussian: Editable 4D Generation through Gaussian Skeletonization}

\author{Lifan Wu \quad Ruijie Zhu \quad Yubo Ai \quad Tianzhu Zhang\\
University of Science and Technology of China\\
{\tt\small \{wusar, ruijiezhu, erebai\}@mail.ustc.edu.cn, tzzhang@ustc.edu.cn}
}

\maketitle

\begin{figure*}[!t]
    \centering
    \includegraphics[width=\linewidth]{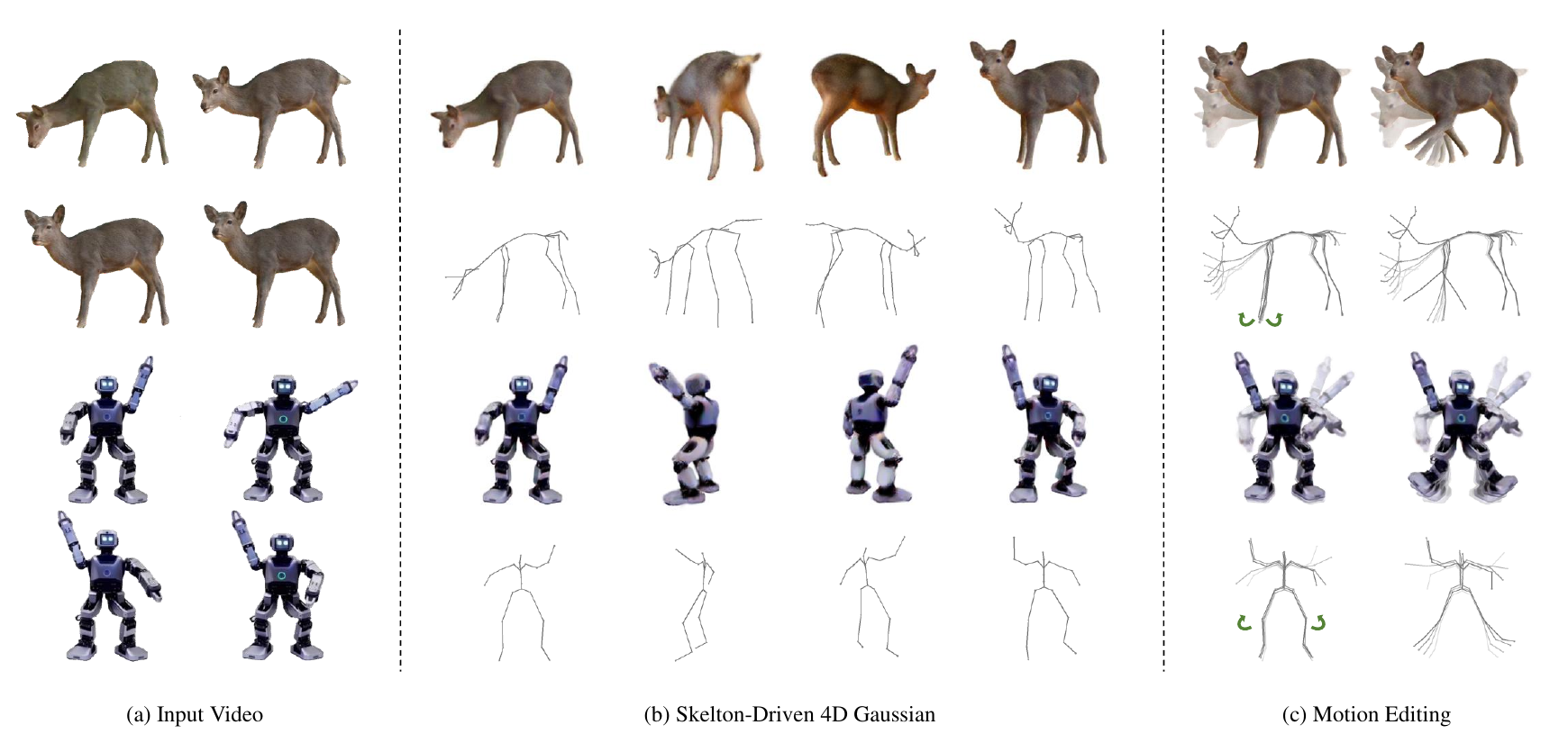}
    \caption{Given (a) an input monocular video, we propose a novel 4D generation method \textbf{SkeletonGaussian} which uses (b) a skeleton to drive the motion of 4D Gaussian model. SkeletonGaussian enables (c) direct motion editing through the skeleton's explicit motion representation, allowing users to adjust skeleton poses to modify the motion of the objects directly.}
    \label{fig:illustrated_pipeline}
\end{figure*}

\input{body/abstract.tex}


\input{body/introduction.tex}
\input{body/relatedwork.tex}

\input{body/method.tex}

\input{body/experiments.tex}

\input{body/summary.tex}


{\small
\bibliographystyle{cvm}
\bibliography{main}
}

\input{body/X_suppl}

\end{document}

%% file: body/abstract.tex
\begin{abstract}
4D generation has made remarkable progress in synthesizing dynamic 3D objects from input text, images, or videos. However, existing methods often represent motion as an implicit deformation field, which limits direct control and editability. To address this, we propose SkeletonGaussian, a novel framework for generating editable, dynamic 3D Gaussians from monocular video input. Our approach introduces a hierarchical, articulated representation that decomposes motion into sparse, rigid motion explicitly driven by a skeleton and fine-grained, non-rigid motion. Concretely, we extract a robust skeleton and drive rigid motion via linear blend skinning, followed by a hexplane-based refinement for non-rigid deformations—enhancing interpretability and editability. Experimental results show that SkeletonGaussian surpasses existing methods in generation quality while enabling intuitive motion editing, establishing a new paradigm for editable 4D generation.
Project page: \url{https://wusar.github.io/projects/skeletongaussian/}
\end{abstract}

%% file: body/introduction.tex
\section{Introduction}

Dynamic 3D generation, also referred to as 4D generation, aims to create dynamic 3D objects from input text, images, or videos. It has become a prominent research area, expanding creative possibilities in fields such as animation, game design, autonomous driving, and film production. In this paper, we focus on generating editable dynamic 3D Gaussian models~\cite{kerbl20233d, wu20244d, yang2024deformable} from monocular video input.

Recent advancements in text-to-3D generation~\cite{lin2023magic3d, poole2022dreamfusion, wang2023score, yu2023points} and image-to-3D synthesis~\cite{lin2023consistent123, liu2024one, qian2023magic123, sun2023dreamcraft3d, tang2023dreamgaussian} have enhanced the creation of diverse 3D objects. Building upon these developments, novel techniques~\cite{kerbl20233d, wu20244d, yang2024deformable} have emerged in the field of 4D generation. These methods leverage Score Distillation Sampling (SDS) loss, derived from diffusion model priors~\cite{xing2023dynamicrafter, shi2023zero123++, liu2023zero, li2024dreammesh4d}, to optimize 4D object representation models. Depending on the type of 4D model used, existing methods can be categorized into three main classes: dynamic Neural Radiance Field (NeRF) generation~\cite{jiang2023consistent4d, mildenhall2021nerf, yuan2022nerf, pons2021d, park2021nerfies, park2021hypernerf}, dynamic 3D Gaussian generation~\cite{pan2024fast}, and dynamic mesh generation~\cite{li2024dreammesh4d}.

Despite these advances, current 4D object representation methods typically model motion as an implicit deformation field~\cite{cao2023hexplane}, which limits direct control and editability. Editing deformation fields~\cite{cao2023hexplane} in 4D models often requires retraining the deformation field, making the process time-consuming and lacking real-time feedback. Moreover, the parameter requirements of the deformation method grow quadratically with time, making it challenging to apply this motion modeling approach to long-duration sequences. Additionally, implicit deformation representations are difficult to convert into standard skeleton or pose data, which obstructs seamless integration with widely used animation tools and pipelines (e.g., Blender~\cite{blender}). Together, these limitations hinder the adoption of practical motion generation workflows.

To tackle these challenges, we aim to develop a high-quality 4D generation workflow that not only produces superior 4D results but also facilitates real-time motion editing. Inspired by recent advances in human reconstruction~\cite{pokhariya2024manus, qian20243dgs, hu2024gaussianavatar, kocabas2024hugs}, which integrate the SMPL model~\cite{loper2023smpl} into 4D Gaussian modeling, we introduce \textbf{SkeletonGaussian}. SkeletonGaussian is an innovative framework for editable 4D generation through Gaussian skeletonization. This framework introduces a lightweight, hierarchical articulated motion representation technique that captures motion details across multiple levels. Therefore, it enables efficient and high-quality 4D generation, while providing flexible editing capabilities.

SkeletonGaussian integrates linear blend skinning (LBS) and skeleton-driven articulated motion representations into 4D generation tasks. It decomposes object motion into two components: sparse rigid deformation, driven by the skeleton, and fine non-rigid deformation, which captures intricate motion details such as wrinkles in clothing and skin. Our 4D generation pipeline consists of three stages: static 3D Gaussian generation, rigid motion modeling, and non-rigid motion refinement. We adopt UniRig~\cite{zhang2025unirig} as the default skeleton extractor for robust, category-agnostic rigging, while using Coverage Axis++ as an ablation baseline. By leveraging hierarchical motion structures, SkeletonGaussian effectively captures complex motion dynamics, particularly in scenarios involving substantial transformations and intricate deformations. Moreover, SkeletonGaussian enables users to directly modify motion by editing the skeleton. It seamlessly integrates into existing 3D animation workflows, allowing for real-time motion adjustments without the need for computationally expensive optimization. The skeletal structure encodes the object's physical topology, eliminating the need for auxiliary constraints, such as ARAP loss. Meanwhile, the explicit skeleton deformation method is highly parameter-efficient. The number of learnable pose parameters grows linearly with joints and time ($\mathcal{O}(B\times T)$), reducing memory and training time compared to dense deformation fields. To validate the effectiveness of our method, we conduct quantitative and qualitative experiments using the Consistent4D~\cite{jiang2023consistent4d} dataset. Our contributions are summarized as follows:

\begin{itemize}  
    \item We propose \textbf{SkeletonGaussian}, a skeleton-driven dynamic 3D Gaussian framework for motion modeling in generative tasks. By employing a hierarchical motion representation, SkeletonGaussian enhances motion fidelity while offering interpretable and editable pose controls. In contrast to dense deformation fields, the skeleton-based pose parameterization is more parameter-efficient, thereby reducing both storage demands and training times.
    \item Our explicit skeleton-based representation enables direct, real-time motion editing through the manipulation of skeletal poses. The generated motions can be exported in standard skeleton and pose formats, ensuring seamless integration with animation pipelines such as Blender~\cite{blender}.

\end{itemize}

%% file: body/relatedwork.tex
\section{Related Work}

\noindent\textbf{Skeleton-Based Motion Representations.} Skeleton-based motion representation is crucial in computer vision and graphics due to its manipulability and ability to model detailed object motion. It is extensively used in computer graphics, animation generation, and pose estimation. Linear Blend Skinning (LBS) is a commonly used technique for animating 3D models by applying transformations to a hierarchical skeleton, where the movement of joints influences the deformation of the model's surface, enabling realistic motion and posing~\cite{lewis2023pose}. LBS is extensively used in contemporary 3D animation and motion modeling. In practical applications, models such as the Skinned Multi-Person Linear Model (SMPL)~\cite{loper2023smpl} employ LBS to combine joint articulation with skin meshes, resulting in realistic human motion and deformation. Similarly, FLAME~\cite{li2017learning} extends this method to facial animation, while SMAL~\cite{zuffi20173d} adapts it for animal modeling, demonstrating its versatility across different specialized fields.

\noindent\textbf{3D Skeleton Generation.} The extraction of skeletons from 3D representations, such as meshes or point clouds, is a well-established study area. Traditional methods relied on hand-crafted rules to extract geometric features. Techniques such as Laplacian contraction~\cite{cao2010point} reduce point clouds to their topological structures, facilitating the extraction of key joints and skeletons. Offline approaches \cite{meyer2023cherrypicker, dou2022coverage, wang2024coverage, wu2015deep, li2015q} have further refined this process. Recent methods~\cite{xu2019predicting, lin2021point2skeleton} employ deep neural networks to predict curve-based skeletons. In our system, we adopt \textbf{UniRig}~\cite{zhang2025unirig} as the default skeleton extractor due to its generalizable rigging prior across categories, while also evaluating \textbf{Coverage Axis++}~\cite{wang2024coverage} as a baseline in our ablations (\Cref{sec:ablationstudy-skeleton}). Both extractors are integrated into a unified pipeline with consistent axis correction and scale normalization.

\noindent\textbf{3D Deformation.} In 3D modeling, deformation techniques incorporate deformation fields into static 3D models. These techniques can be classified based on the model's 3D representation: (1) \textbf{Mesh Deformation.} Classical mesh-based methods, such as Laplacian coordinates~\cite{lipman2005laplacian, sorkine2007rigid} and cage-based techniques~\cite{yifan2020neural}, focus on preserving geometric details during transformations, making them suitable for static objects. (2) \textbf{NeRF-Based Deformation.} Recent advancements in dynamic NeRF reconstruction utilize plane decomposition and 4D grids~\cite{cao2023hexplane, fridovich2023k} to achieve dynamic scene reconstruction by deforming canonical NeRF. \revised{(3) \textbf{3D Gaussian Deformation.}} Recent advancements in 3D Gaussian splatting~\cite{kerbl20233d} significantly accelerate rendering. Dynamic 3D Gaussian methods~\cite{luiten2023dynamic, wu20244d, yang2024deformable, zhu2024motiongs, lu2024dn} leverage deformation fields~\cite{cao2023hexplane} to model 3D Gaussian motion. Some approaches, such as SC-GS~\cite{huang2024sc} and BAGS~\cite{zhang2024bags}, use sparse control points to represent 3D deformation. However, these methods rely on hexplane and MLP-based techniques to implicitly model control point movement and deformation, whereas our method explicitly models motion using a skeleton and joint pose parameterization. Recent techniques for 3D Gaussian-based dynamic human motion~\cite{pokhariya2024manus, qian20243dgs, hu2024gaussianavatar, kocabas2024hugs} combine rigid skeletal skinning and non-rigid deformations for precise motion modeling. Our approach draws inspiration from these works, employing a skeleton-based deformation in 3D Gaussian rendering to provide an intuitive and efficient method for editable 4D generation.

\noindent\textbf{3D and 4D Generation.} In 3D generation, DreamFusion~\cite{poole2022dreamfusion} first introduced score distillation sampling (SDS)~\cite{poole2022dreamfusion, wang2023score} loss, which optimizes NeRF~\cite{mildenhall2021nerf} to produce high-quality 3D models. Building on advancements in 3D Gaussian techniques~\cite{kerbl20233d}, DreamGaussian~\cite{tang2023dreamgaussian} leverages Gaussian splatting for 3D generation, significantly improving speed and performance. In 4D generation research, traditional motion generation methods~\cite{harvey2020robust, starke2023motion} are often limited to specific characters or datasets, reducing their applicability across diverse objects. Recently, diffusion model-based approaches~\cite{singer2023text, bahmani20244d, ling2024align, ren2023dreamgaussian4d, zhao2023animate124, jiang2023consistent4d, zeng2024stag4d, yin20234dgen} address these limitations by integrating SDS loss into 4D framework, enabling more versatile and generalized motion generation. Approaches such as SC4D~\cite{wu2024sc4d} introduce control points for motion transfer, enhancing editing flexibility in 4D generation. Additionally, Diffusion4D~\cite{liang2024diffusion4d} and Stable Video 4D~\cite{xie2024sv4d} achieve spatiotemporal consistency through specialized attention layers. STAG4D~\cite{zeng2024stag4d} initializes multi-view images anchored to input video frames, which are then used for multi-view SDS computation. Building on these advancements, our work introduces a novel skeleton-driven 3D Gaussian deformation framework for 4D generation tasks, offering a more intuitive and efficient approach to motion modeling and editing.

%% file: body/method.tex
\begin{figure*}
    \centering
    \includegraphics[width=\linewidth]{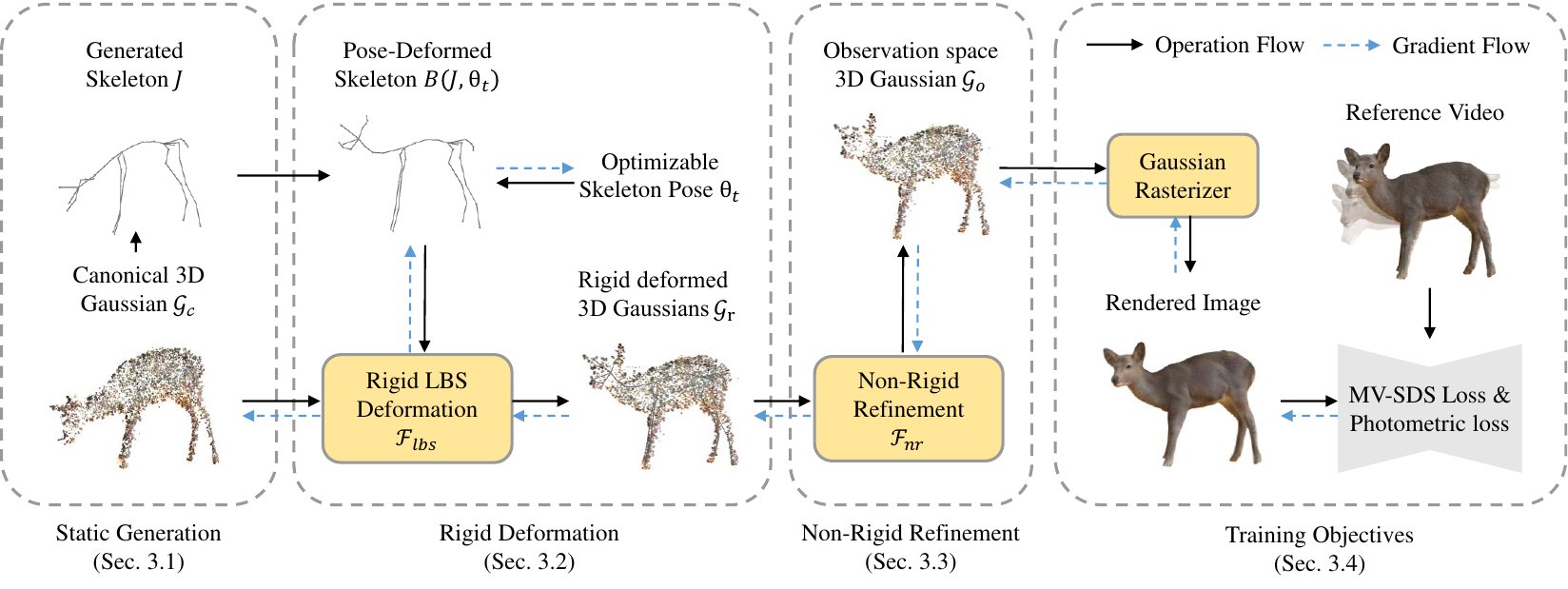}
    \caption{Pipeline of the SkeletonGaussian framework for 4D object generation, divided into three stages: (1) \textbf{Static 3D Object Generation and Skeleton Extraction}: Starting from a frame at the video's midpoint, a static 3D Gaussian model \( \mathcal{G}_{c} \) (\Cref{sec:static_3d_generation}) is generated in canonical space, from which an inherent skeletal structure is subsequently extracted. (2) \textbf{Rigid Motion Modeling}: Using LBS, rigid deformations \( \mathcal{F}_{lbs} \) (\Cref{sec:rigid_lbs_deformation}) under various poses \( \theta_t \) are applied to rigidly deform \( \mathcal{G}_c \) into \( \mathcal{G}_r \). During this stage, the skeleton poses \( \theta_t \) are optimized. (3) \textbf{Non-Rigid Motion Modeling}: To capture fine-grained deformations, a deformation field \( \mathcal{F}_{nr} \) (\Cref{sec:non_rigid_refinement}) refines the motion of the rigidly deformed 3D Gaussian \( \mathcal{G}_r \), transforming it into the observation space Gaussian \( \mathcal{G}_o \). \( \mathcal{F}_{nr} \) comprises a hexplane~\cite{cao2023hexplane} and an MLP. All three stages share the same \textbf{Training Objectives} (\Cref{sec:training_objectives}). A differentiable Gaussian rasterizer renders images of the observation space 3D Gaussian \( \mathcal{G}_o \) from multiple viewpoints, comparing them to the reference video with photometric and MV-SDS losses for backpropagation.} 
    \label{fig:full_pipeline}
\end{figure*}

\section{Method}

As illustrated in \Cref{fig:full_pipeline}, the 4D object generation pipeline of SkeletonGaussian consists of three stages: (1) \textbf{Static 3D Object Generation and Skeleton Extraction} (\Cref{sec:static_3d_generation}): A static 3D object is initially generated using a 3D Gaussian generation method~\cite{kerbl20233d, tang2023dreamgaussian}. Subsequently, an inherent skeletal structure is constructed for the 3D Gaussian model. (2) \textbf{Rigid Motion Modeling} (\Cref{sec:rigid_lbs_deformation}): To capture the primary rigid motion of the object, a skeletal skinning network is employed, and the motion trajectories of each skeletal point are calculated using Forward Kinematics. Linear Blend Skinning (LBS) is then applied to deform the 3D Gaussian model according to these skeletal trajectories. During this stage, the \textbf{skeleton poses} are optimized to match the reference video sequences. (3) \textbf{Non-Rigid Motion Modeling} (\Cref{sec:non_rigid_refinement}): Fine-grained non-rigid motions are represented through a hexplane~\cite{cao2023hexplane} and a deformation MLP. At this stage, the skeletal skinning network remains frozen, while only the 3D Gaussian and the hexplane deformation field are trained to capture finer deformations.

Through these three stages, SkeletonGaussian generates a high-quality 4D object comprising a 3D Gaussian model, a skeletal pose representing the object's rigid motion, and a fine-grained deformation field, achieving high-quality dynamic 3D object generation. The following sections detail each of these training stages.

\input{body/method1}

\input{body/method2}

\input{body/method3}

\input{body/method4}

%% file: body/method1.tex
\subsection{Static 3D Gaussian and Skeleton Generation}
\label{sec:static_3d_generation}
To generate a static 3D Gaussian and its corresponding skeletal structure, we select the middle frame of the video as the reference frame for constructing the initial static 3D Gaussian model \( \mathcal{G}_c \) in canonical space:

\begin{equation}
\mathcal{G}_c = \{\mathbf{p}_c,\mathbf{q}_c, \mathbf{s}, \sigma, \mathbf{c}\},
\end{equation}

\noindent where \(\mathbf{p}_c\), \( \mathbf{q}_c \), \( \mathbf{s} \), \( \sigma \), and \( \mathbf{c} \) represent the position, quaternions, scale, opacity, and spherical harmonics coefficients of the 3D Gaussian in canonical space, respectively. The middle frame is chosen as the static reference because it minimizes the motion discrepancy with all other frames, thereby reducing task complexity. The static 3D Gaussian model is trained using both the multi-view SDS loss and the photometric consistency loss. Further details are provided in \Cref{sec:training_objectives}.

\noindent\textbf{Skeleton Generation.} To generate the kinematic tree structure of the skeleton joints \(\mathbf{J}\), the mesh structure of the static object is first extracted from the static 3D Gaussian using occupation fields and the marching cubes algorithm~\cite{lorensen1998marching}. We adopt a robust rigging pipeline built on UniRig~\cite{zhang2025unirig} (default in our system), which predicts joint candidates and their connectivity to form an articulated skeleton for general objects. In practice, we support two invocation modes to enhance reproducibility and portability across environments: (1) an \emph{internal} Python inference path, and (2) an \emph{external} script path that caches results on disk and can be launched from sandboxed environments. Prior to building forward kinematics (FK), we apply standard preprocessing to ensure consistent coordinate conventions across extractors.

Based on the extracted skeletal points, we construct a kinematic tree by computing a Minimum Spanning Tree (MST) over candidate joints, and we preserve joint identifiers when available from the extractor to maintain consistency with rigging conventions. The kinematic tree provides a compact structural abstraction of the 3D Gaussian and serves as the control scaffold for subsequent motion generation.

%% file: body/method2.tex
\subsection{3D Gaussian Rigid Deformation}

\label{sec:rigid_lbs_deformation}

To model the primary motion of the 3D Gaussian, we use LBS to apply rigid deformation to the canonical 3D Gaussian \( \mathcal{G}_c \), denoted as \( \mathcal{F}_{lbs} \). Let \( \mathbf{J} = \{\mathbf{J}_b\}_{b=1}^B \) represent the set of static joint positions of the skeleton, and let \( \theta_t \) represent the skeleton's pose at a specific time \( t \). Under these conditions, the corresponding rigidly deformed 3D Gaussian \( \mathcal{G}_r \) is computed as follows:

\begin{equation}
\mathcal{G}_r = \mathcal{F}_{lbs}(\mathcal{G}_c; \mathbf{J}, \theta_t).
\end{equation}
\revised{Note that while LBS results in non-rigid deformation, we term this ``rigid deformation'' to highlight it is driven by rigid skeletal joints, distinguishing it from the subsequent non-rigid refinement.}

\noindent\textbf{Rigid Position and Rotation Transform.} The deformed 3D Gaussian point \(\mathcal{G}_r^i\) is computed by applying a transformation matrix \(\mathbf{T}_i\) to the 3D Gaussian point \(\mathcal{G}_c^i\). \(\mathbf{T}_i\) is a weighted sum of the transformation matrices \( \mathbf{B}_k(\mathbf{J}, \theta_t) \) corresponding to the nearest \(K\) skeletal joints, with weights \(w_{k,i}\):
\begin{equation}
\mathbf{T}_i = \sum_{k=1}^{K} w_{k,i} \mathbf{B}_k(\mathbf{J}, \theta_t).
\end{equation}
The transformed position \(\mathbf{p}_r^i\) and the rotation \(\mathbf{q}_r^i\) of the deformed 3D Gaussian \(\mathcal{G}_c^i\) are calculated as follows:

\begin{equation}
\mathbf{p}_r^i = \mathbf{T}_i\mathbf{p}_c^i + \mathbf{t}_o, \quad \mathbf{q}_r^i = \mathbf{T}_{i \left(1:3,1:3\right)} \cdot \mathbf{q}_c^i,
\end{equation}

\noindent where \(\mathbf{T}_{1:3,1:3}\) refers to the rotational component extracted from the transformation matrix \(\mathbf{T}_i\).
The term \( \mathbf{t}_o \) denotes the global translation of the root joint at time \(t\).

\begin{figure*}
    \centering
    \includegraphics[width=\linewidth]{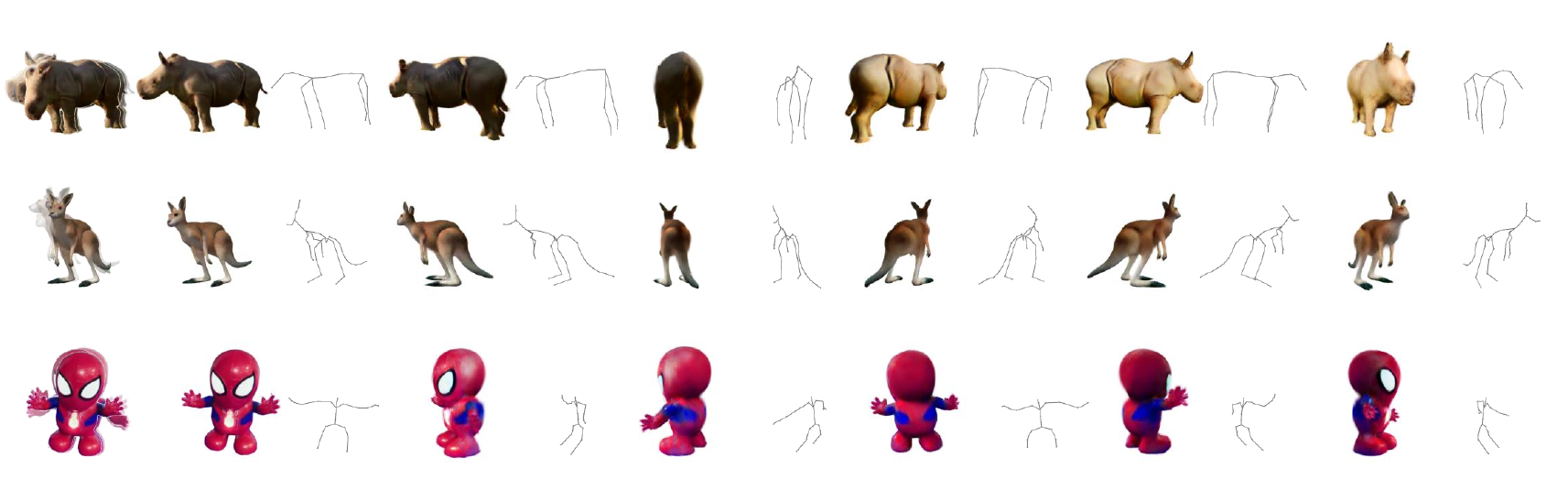}
    \caption{Visualizing 4D Object Motion with Skeleton Poses. We present generated 4D object motion and its corresponding skeleton poses, where the viewpoint rotates from left to right, and time progresses linearly from left to right.}
    \label{fig:rotation_view}
\end{figure*}

\noindent\textbf{Forward Kinematics.} To compute skeletal animations and joint motions, we use the forward kinematics approach. Forward kinematics determines each joint's transformation by recursively accumulating the transformations of its ancestor joints. It relies on a hierarchical skeleton tree structure, where the transformation matrix \( \mathbf{B}_k(\mathbf{J}, \theta_t) \) for each joint \( k \) is obtained by multiplying the local transformations \(\theta_t^j\) of all its ancestor joints \( j \in A(k) \):

\begin{equation}
\mathbf{B}_k(\mathbf{J}, \theta_t) = \prod_{j \in A(k)} \theta_{t,j}.
\end{equation}

\noindent\textbf{Skeletal Skinning Weights.} To compute the skinning weights \( w_{k, i} \) for each Gaussian point \( i \) relative to its surrounding skeleton joints, we apply the K-nearest neighbors (KNN) algorithm to identify the \( K \) nearest points. The weights are determined using inverse distance weighting, where the weight is inversely proportional to the distance \( d_{k, i} \) between point \( i \) and skeleton joint \( k \):

\begin{equation}
w_{k, i} = \frac{\frac{1}{d_{k, i}}}{\sum_{k=1}^{K} \frac{1}{d_{k, i}}}.
\end{equation}
\revised{We employ fixed inverse-distance KNN weights due to their simplicity and lack of training requirements. In future work, we plan to explore learning-based skinning weight fields to further improve deformation quality.}

%% file: body/method3.tex
\label{sec:skeleton_pose_training}
\noindent\textbf{\revised{Skeletal Pose Smoothness.}} The skeletal pose is represented by a tensor \( \theta \in \mathbb{R}^{T \times \revised{B} \times 4} \), where \( T \) is the number of frames, \( \revised{B} \) is the number of joints, and each entry \( \theta_{t,k} \) represents a 4D quaternion encoding the rotation of the \( k \)-th joint at the \( t \)-th frame. Additionally, we introduce a variable \( \mathbf{t}_o \) to record the global translation of the root joint. Directly optimizing skeleton poses can lead to overfitting, causing the model to capture noise from the training data and produce jitter in the generated motion. To mitigate this problem, we employ window smoothing to the skeleton poses, which uses a sliding window of size \(2w + 1\) during training to smooth the motion across \(w\) neighboring frames. For the local skeletal pose \(\theta_t\) at each time step \(t\), we compute the average pose \(\bar{\theta_t}\) by averaging across frame \(t\) and its neighbors. This smoothed value \(\bar{\theta_t}\) is then used as input for LBS deformations. For simplicity, we use \( \theta_t \) instead of \( \bar{\theta_t} \) elsewhere. The formula is:

\begin{equation}
\bar{\theta_t} = \frac{1}{2w + 1} \sum_{i=-w}^{w} \theta_{t+i}.
\end{equation}

%% file: body/method4.tex
\begin{figure}
    \centering
    \includegraphics[width=\linewidth]{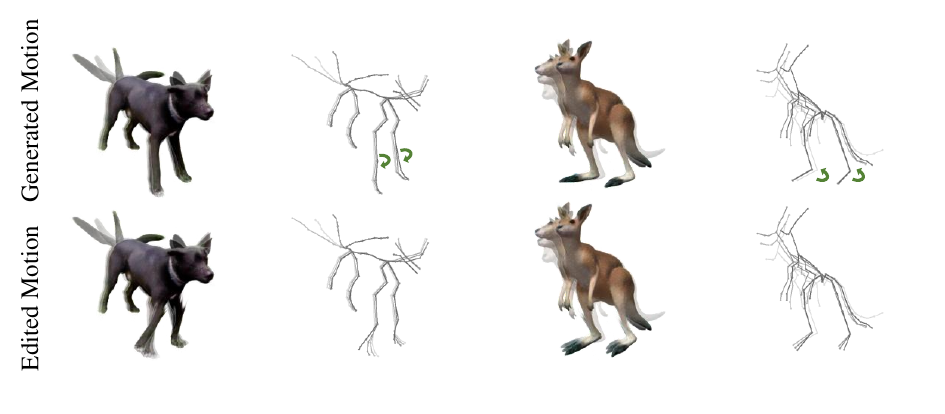}
    \caption{Editing Generated Motion. We visualize the generated motion (top) and edited motion sequence (bottom). Users can directly adjust the skeleton poses of specific joints at different times to edit the object's motion.}
    \label{fig:editing}
\end{figure}

\subsection{Non-Rigid Refinement}
\label{sec:non_rigid_refinement}
LBS effectively captures global object motion but struggles to represent detailed motions, such as clothing wrinkles, due to the limited number of skeletal joints. To overcome this limitation, we propose a non-rigid deformation method to capture these detailed motions. Our approach employs a hexplane-based 4D representation to refine the motion of static 3D Gaussians. Specifically, we integrate a hexplane and an MLP to regress displacement, rotation, and scale changes of the 3D Gaussian. The non-rigid deformation function \(\mathcal{F}_{nr}\) which transforms the rigidly deformed 3D Gaussian \(\mathcal{G}_r\) into the observation space 3D Gaussian \(\mathcal{G}_o\) is given by:
\begin{equation}
\mathcal{G}_o = \mathcal{F}_{nr}(\mathcal{G}_r).
\end{equation}
The observation space 3D Gaussian \(\mathcal{G}_o\) is then rendered through Gaussian rasterization. During this refinement field training stage, the skeletal skinning network is frozen, and only the 3D Gaussian and hexplane deformation field are trained to capture fine-grained deformations.

\subsection{Training Objectives}
\label{sec:training_objectives}
Our objective is to generate a 4D Gaussian representation of the target object from an input video sequence by optimizing the static 3D Gaussian model, the skeleton poses \(\theta_t\), and the refinement field parameters \(\mathcal{F}_{nr}\). We begin by applying the multi-view diffusion model Zero123++~\cite{shi2023zero123++} to generate multi-view sequences \( I_{\text{anchor}}^t \) from the video inputs. These sequences act as spatiotemporal anchors, which are later used to compute the MV-SDS loss. The 3D Gaussian is then projected onto the screen to produce output images, which are compared with the anchor images \(I_{\text{anchor}}^t\) using the multi-view Score Distillation Sampling (SDS) loss \(\mathcal{L}_{MV-SDS}\) from Zero123~\cite{liu2023zero}. In addition, the loss function includes a reconstruction loss \(\mathcal{L}_{rec}\) and a foreground masking loss \(\mathcal{L}_{mask}\) between the reference image \(I_t^{ref}\) and the front-view rendered image. A regularization loss \(\mathcal{L}_{reg}\) is also applied to the deformation field \(\mathcal{F}_{nr}\) to enforce temporal smoothness in the motion. For a detailed explanation of the loss function, please refer to the Appendix \Cref{sec:loss_functions}. The final optimization objective is given by:

\begin{equation}
\mathcal{L} = \mathcal{L}_{MV-SDS} + \lambda_1 \mathcal{L}_{rec} + \lambda_2 \mathcal{L}_{mask} + \lambda_3 \mathcal{L}_{reg}.
\end{equation}

\subsection{Generated Motion Editing}

SkeletonGaussian provides an efficient approach for editing generated motion through its sparse, explicit skeleton-based representation. Users can intuitively adjust the motion of skeletal points by modifying poses at specific time steps, thereby altering the entire motion trajectory. We develop a GUI that simplifies the process of motion editing. This method also aligns with current motion modeling techniques in computer graphics, enabling users to modify skeletal movements in popular 3D editors such as Blender~\cite{blender}. Additionally, the hierarchical structure of the skeleton tree facilitates hierarchical motion editing, where adjustments to a parent node automatically propagate to its child nodes. Motion editing is illustrated in \Cref{fig:editing}.

%% file: body/experiments.tex
\begin{figure*}
    \centering
    \includegraphics[width=\linewidth]{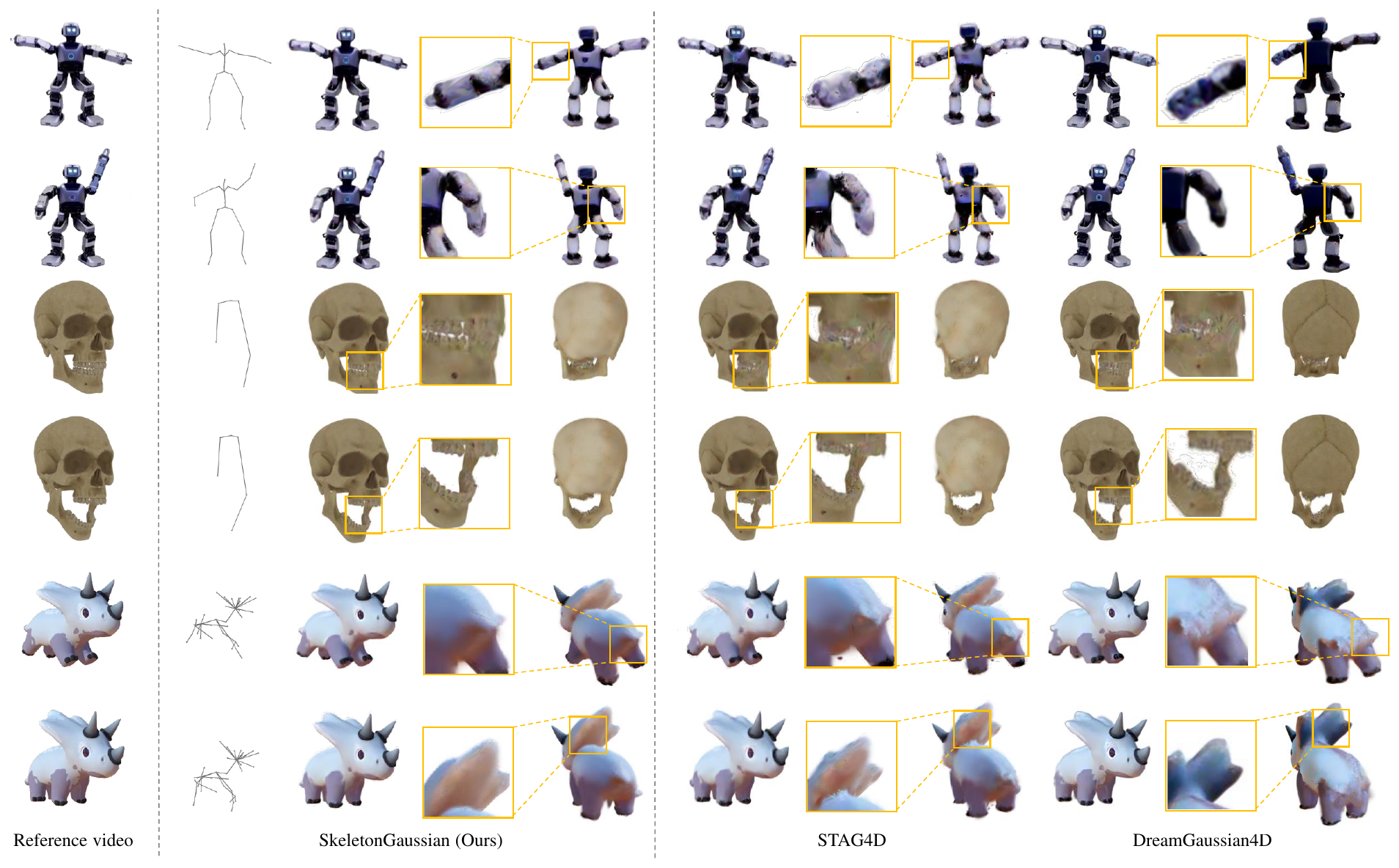}
    \caption{Qualitative Comparisons. We compare our method with STAG4D~\cite{zeng2024stag4d} and DreamGaussian4D~\cite{ren2023dreamgaussian4d}. For each instance, we render two viewpoints at two time steps. We also visualize the skeleton poses of SkeletonGaussian.}
    \label{fig:comparison_results}
\end{figure*}

\section{Experiments}

\subsection{Experiment Setup}
\noindent\textbf{Implementation Details.} (1) \textbf{Static Stage}: We generate six anchor view videos \({I_t^i}\) (\(i \in \{1...6\}\)) using Zero123++~\cite{shi2023zero123++} from the input monocular video \(I_t^{ref}\). The SDS loss is computed using Zero-1-to-3~\cite{liu2023zero}. 10000 3D Gaussian points are randomly initialized within a spherical canonical space. This stage is trained for 1500 steps to produce a static 3D Gaussian. Subsequently, a skeleton is generated using \textbf{UniRig}~\cite{zhang2025unirig} (default). We support both an internal Python path and an external cached script path for cross-environment execution. (2) \textbf{Skeleton Training Stage}: Skeleton poses are trained for 2500 steps. A smoothing window of size three is applied to the skeleton poses. (3) \textbf{Non-Rigid Motion Refinement}: A hexplane and a deformation MLP are trained for 7000 steps to capture the fine-grained motion. The detailed implementation and hyperparameters are provided in Appendix \Cref{sec:implementation_details}. The entire training process takes approximately 1 hour on an RTX 3090 GPU, and the rendering process can be performed at 150 FPS in real time.

\noindent\textbf{Evaluation Dataset.} To fairly evaluate our method against the baselines, we use the Consistent4D dataset~\cite{jiang2023consistent4d}, which includes 4D animation assets from Sketchfab~\cite{sketchfab} for further animation assessment. The dataset comprises 12 synthetic and 12 real-world videos, each captured with a static vertically aligned camera focused on dynamic objects. Each video contains 32 frames over approximately 2 seconds.
\label{sec:evaluation_metrics}

\noindent\textbf{Evaluation Metrics.} We evaluate the quality of generated 4D videos based on their alignment with reference videos, spatio-temporal consistency, and motion fidelity. For each test object and method, we use a frontal view video as input to generate a corresponding dynamic 3D model, rendering four videos from azimuth angles of 75°, 15°, 105°, and 195° at a 0° elevation. These rendered videos are compared with the ground-truth videos in the dataset to evaluate the generation quality. Our evaluation metrics include CLIP~\cite{radford2021learning}, LPIPS~\cite{zhang2018unreasonable}, and FVD~\cite{unterthiner2018towards} for Video-to-4D evaluation. CLIP and LPIPS evaluate the semantic and perceptual similarities between generated and real images, while FVD computes frame quality and temporal consistency. Since DreamGaussian4D generates videos with only 16 frames, we use the FVD-16 score, which computes the FVD based on the first 16 frames.

\noindent\textbf{Baselines.} We compare SkeletonGaussian with several recent 4D generation methods capable of generating multi-view videos from a single-view video input, including Consistent4D~\cite{jiang2023consistent4d}, STAG4D~\cite{zeng2024stag4d}, 4DGen~\cite{yin20234dgen}, and DreamGaussian4D~\cite{ren2023dreamgaussian4d}. All baselines are evaluated on the Consistent4D dataset using their official code and configurations. Quantitative and qualitative comparisons are presented in \Cref{fig:comparison_results} and \Cref{tab:quantitative-comparison}.

\begin{table}
\centering
\caption{\textbf{Quantitative Evaluation of 4D Generation on the Consistent4D Dataset.} SkeletonGaussian outperforms in both image quality and video frame consistency. }
\label{tab:quantitative-comparison}
\resizebox{\linewidth}{!}{
\begin{tabular}{@{}cccccc@{}}
\toprule
Method  & CLIP $\uparrow$ & LPIPS $\downarrow$ & FVD $\downarrow$ \\ 
\midrule
Consistent4D~\cite{jiang2023consistent4d} & 0.877 & 0.161 & 1518.5 \\
DreamGaussian4D~\cite{ren2023dreamgaussian4d} & 0.913 & 0.143 &  994.11 \\
4DGen(16 frames)~\cite{yin20234dgen} & 0.909 & 0.137 & 913.10 \\
STAG4D~\cite{zeng2024stag4d}& 0.909 & 0.126 & 992.2 \\
\midrule
\textbf{SkeletonGaussian (Ours)} & \textbf{0.923} & \textbf{0.125} & \textbf{847.8}  \\
\bottomrule
\end{tabular}
}
\end{table}

\subsection{Comparisons}

\noindent\textbf{Quantitative Comparisons.} As shown in \Cref{tab:quantitative-comparison}, our method outperforms STAG4D, DreamGaussian4D, and 4DGen on the Consistent4D~\cite{jiang2023consistent4d} dataset in terms of reference view alignment (LPIPS, CLIP), indicating that our approach generates more realistic images. Furthermore, our method achieves the lowest FVD score, demonstrating that our generated videos exhibit fewer temporal artifacts and better match real-world footage. These results highlight the effectiveness of SkeletonGaussian.

\noindent\textbf{Qualitative Comparison.} We compare the 4D outputs generated by our method with STAG4D and DreamGaussian4D in \Cref{fig:comparison_results}. For each video, we render the 4D results at two timestamps and from two perspectives: one from the front and the other from the back. Additionally, we visualize the skeletons of our method. Our approach achieves high-fidelity reconstruction with stable geometry and consistent texture. The results maintain fine details across frames, demonstrating robustness in both spatial and temporal aspects.

\noindent\textbf{User Study.} To validate our method, we conduct user studies to evaluate multi-view video synthesis and 4D outputs. We select 20 real-world and synthetic videos from the Objaverse~\cite{objaverse} and Consistent4D~\cite{jiang2023consistent4d} datasets. Participants compare results from three methods (SkeletonGaussian and three baselines) based on a novel camera view, choosing the most stable, realistic, and reference-like video. SkeletonGaussian is preferred by 32.5\%, followed by STAG4D (27.5\%), 4DGen (22.5\%), and DreamGaussian4D (17.5\%).

\begin{figure*}
    \centering
    \includegraphics[width=\linewidth]{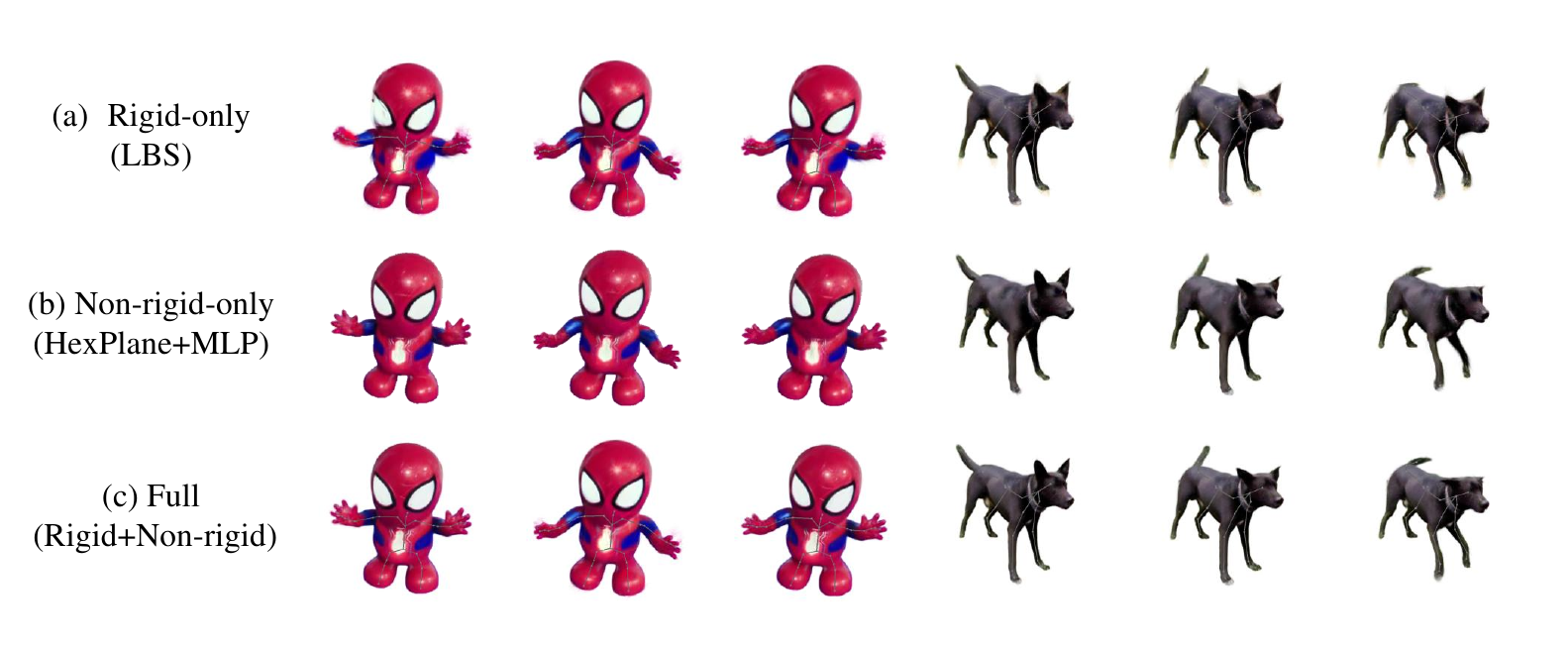}
    \caption{Qualitative evaluations of the ablation study. We visualize the skeleton poses and the objects at different time steps.}
    \label{fig:ablationstudy_graph}
\end{figure*}

\subsection{Ablation Studies}
\label{sec:ablationstudy}
\label{sec:ablationstudy-skeleton}

In this section, we evaluate the effectiveness of various motion modeling methods by analyzing the quality of the generated 4D Gaussians through a series of ablation studies. We assess the quality, memory requirements, and training time of different motion modeling approaches, providing quantitative comparisons in \Cref{tab:ablation_study} and qualitative comparisons in \Cref{fig:ablationstudy_graph}.

\noindent\textbf{Rigid-only (LBS).} Using only rigid LBS preserves articulated structure but underfits fine non-rigid motion (see \Cref{fig:ablationstudy_graph}). Thanks to the compact pose parameterization that scales as $\mathcal{O}(B\!\times\!3\!\times\!T)$, it achieves the smallest deformation-module VRAM and the shortest training time in \Cref{tab:ablation_study}. In our setting with $B\!\approx\!30$ and $T\!=\!32$, this amounts to $30\!\times\!32\!\times\!3$ scalars, i.e., only storing joint rotation angles; the rigid stage takes about 1000 steps (\(~0.2\,\mathrm{h}\)).

\noindent\textbf{Non-rigid-only (HexPlane+MLP).} The non-rigid field captures detailed deformations but lacks an articulated prior, reducing temporal stability. Its parameter count scales with grid/plane resolutions and MLP widths, leading to the largest deformation-module VRAM and the longest training time; empirically the memory cost grows roughly $\mathcal{O}(T^2)$ with sequence length, making long sequences hard to optimize. On Consistent4D with $T\!=\!32$, we measure the deformation module VRAM at 136.40 MiB, and the deformation stage takes about 8000 steps (\(~1.5\,\mathrm{h}\)). Quantitatively, it can slightly improve per-frame fidelity (LPIPS/CLIP) but still trails the Full model on overall temporal quality.

\noindent The \textbf{Full (Rigid+Non-rigid)} model combines both stages; its deformation-module VRAM is close to the non-rigid variant with a negligible skeleton overhead, and the total training time is roughly the sum of the two stages (\(~1.7\,\mathrm{h}\)).

\begin{table*}[t]
\centering
\caption{Quantitative ablation on motion modeling. Metrics include CLIP/LPIPS/FVD and efficiency (VRAM in MiB and training time in minutes) measured for the deformation module only, excluding the static 3D Gaussian. We compare Rigid-only (LBS), Non-rigid-only (HexPlane+MLP), and Full (Rigid+Non-rigid).}
\label{tab:ablation_study}
\resizebox{0.8\textwidth}{!}{
\begin{tabular}{@{}lccccc@{}}
\toprule
Method  & CLIP $\uparrow$ & LPIPS $\downarrow$ & FVD $\downarrow$ & VRAM (MiB) $\downarrow$ & Train Time (min) $\downarrow$ \\
\midrule
Rigid-only (LBS) & 0.901 & 0.135 & 1012.6 & 0.01 & 12 \\
Non-rigid-only (HexPlane+MLP) & 0.909 & 0.126 & 992.2 & 136.40 & 90 \\
\midrule
\textbf{Full (Rigid+Non-rigid)} & \textbf{0.923} & \textbf{0.125} & \textbf{847.8} & 136.41 & 102 \\
\bottomrule
\end{tabular}
}
\end{table*}

\noindent\textbf{Pose Smoothness.} The pose-smoothness regularizer improves the temporal continuity of articulated poses, yielding smoother motion and fewer jitters. Quantitatively, removing it degrades LPIPS/FVD, as summarized in \Cref{tab:ablation_pose_smooth}.

\noindent\textbf{Skeleton Extractor (Coverage Axis++ vs UniRig).} We also ablate the skeleton extractor under the same pipeline. Coverage Axis++~\cite{wang2024coverage} selects skeletal points via coverage heuristics and connects them via a Minimum Spanning Tree, whereas \textbf{UniRig}~\cite{zhang2025unirig} provides a stronger, category-agnostic rigging prior with joint proposals and connectivity that we further regularize via FK. Replacing UniRig with Coverage Axis++ (row ``Using Coverage Axis++'' in \Cref{tab:ablation_pose_smooth}) weakens temporal stability and increases artifacts. Qualitatively, \Cref{fig:ablationstudy_graph} shows more stable and semantically aligned joints with UniRig, which improves control and reduces topological errors.

\noindent\textbf{Impact of Initial Frame Selection.} We further analyze how the choice of the initial reference frame affects the generation results. Our method typically uses the first frame as the canonical reference. Experiments indicate that selecting a frame with clear visibility and a neutral pose contributes to a more accurate canonical 3D Gaussian initialization. However, our skeleton-driven deformation mechanism provides strong geometric priors, making the system relatively robust to the initial frame selection. Even when initialized from frames with partial self-occlusions, the method can recover plausible motion dynamics through the subsequent rigid and non-rigid optimization stages. Quantitative results are shown in \Cref{tab:ablation_initial_frame}.

\begin{table}[h]
\centering
\caption{Ablation study on the impact of initial frame selection. We compare selecting the first frame (Frame 0), the middle frame (Frame 15), and a random frame as the initialization for the 3D Gaussian field. The results show consistent performance across different initial frames.}
\label{tab:ablation_initial_frame}
\begin{tabular}{@{}lccc@{}}
\toprule
Initial Frame Selection & CLIP $\uparrow$ & LPIPS $\downarrow$ & FVD $\downarrow$ \\
\midrule
Frame 0 (First Frame) & 0.921 & 0.126 & 851.2 \\
Frame 15 (Middle Frame) & \textbf{0.923} & \textbf{0.125} & \textbf{847.8} \\
Random Frame & 0.919 & 0.128 & 858.5 \\
\bottomrule
\end{tabular}
\end{table}

\begin{table}[t]
\centering
\caption{Compact ablations on pose smoothness and skeleton extractor. ``Coverage Axis++'' swaps UniRig in the Full (Rigid+Non-rigid) model; ``w/o Pose Smoothness'' drops the pose-smoothness regularizer; ``Full'' uses UniRig with pose smoothness.}
\label{tab:ablation_pose_smooth}
\begin{tabular}{@{}lccc@{}}
\toprule
Method & CLIP $\uparrow$ & LPIPS $\downarrow$ & FVD $\downarrow$ \\
\midrule
Using Coverage Axis++ & 0.918 & 0.128 & 890.4 \\
w/o Pose Smoothness & 0.906 & 0.131 & 1034.3 \\
Full (Rigid+Non-rigid) & \textbf{0.923} & \textbf{0.125} & \textbf{847.8} \\
\bottomrule
\end{tabular}
\end{table}

%% file: body/summary.tex
\section{Discussion}
\label{sec:discussion}

\noindent\textbf{Limitations and Future Directions.} We observe that incorrect skeleton retrieval can degrade the quality of the generated results. Specifically, in some cases, severe topological errors in skeleton extraction can diminish the quality of the generated results. Additionally, there may be cases where objects do not have a clear skeleton structure, and our method performs poorly in these situations. The hexplane deformation field demonstrates error compensation capabilities for mild skeletal inaccuracies, helping to address this issue. Furthermore, we are developing an adaptive skeleton error-correcting mechanism that dynamically adjusts the skeleton structure during training. Please refer to Appendix \Cref{sec:discussion_on_failed_cases} for a detailed analysis of failure cases.

Currently, our method does not support multi-object motion, thus limiting its applicability in scenarios involving multiple objects. Future work could address this limitation by incorporating independent skeletons for each object. Additionally, we are developing the integration of predefined skeleton templates, such as SMPL~\cite{loper2023smpl}, to initialize the 3D Gaussian and skeleton structure using vertex and joint positions. We also successfully integrate the human pose estimation method ViTPose~\cite{xu2022vitpose} into SkeletonGaussian to initialize the skeleton poses. This integration is expected to improve the accuracy and quality of 4D motion generation significantly. Furthermore, our approach can seamlessly integrate with skeleton-controlled video generation techniques, such as ControlNet~\cite{zhang2023adding, guo2023animatediff, hu2024animate}. Using 3D skeletons as conditional inputs for diffusion models in 2D images opens new possibilities for 4D generation. Additionally, enhanced skeletal control provides a novel representation of motion, which could be applied to motion-tracking tasks.

\section{Conclusion}

This paper introduces SkeletonGaussian, a framework for generating editable 4D Gaussian-based models from monocular video. By explicitly decomposing motion into rigid skeletal movements and fine-grained non-rigid details, this framework improves control and interpretability in 4D Gaussian modeling. SkeletonGaussian operates in three phases: constructing a static 3D Gaussian model, modeling rigid motion through skeletal LBS, and refining non-rigid motion using a hexplane-based deformation field. This hierarchical structure enables intuitive motion editing by adjusting skeleton poses and aligning seamlessly with standard animation workflows. Experimental results demonstrate that SkeletonGaussian delivers superior quality over existing methods, offering a new paradigm for editable 4D motion generation.

%% file: body/X_suppl.tex
\clearpage
\setcounter{page}{1}

\newcommand{\maketitlesupplementary}{
    \twocolumn[
    \centering
    \Large
    \textbf{Supplementary Material for \\ SkeletonGaussian: Editable 4D Generation through Gaussian Skeletonization}
    \vspace{0.5cm}
    ]
}
\maketitlesupplementary

\begin{figure*}[hb]
    \centering
    \includegraphics[width=\linewidth]{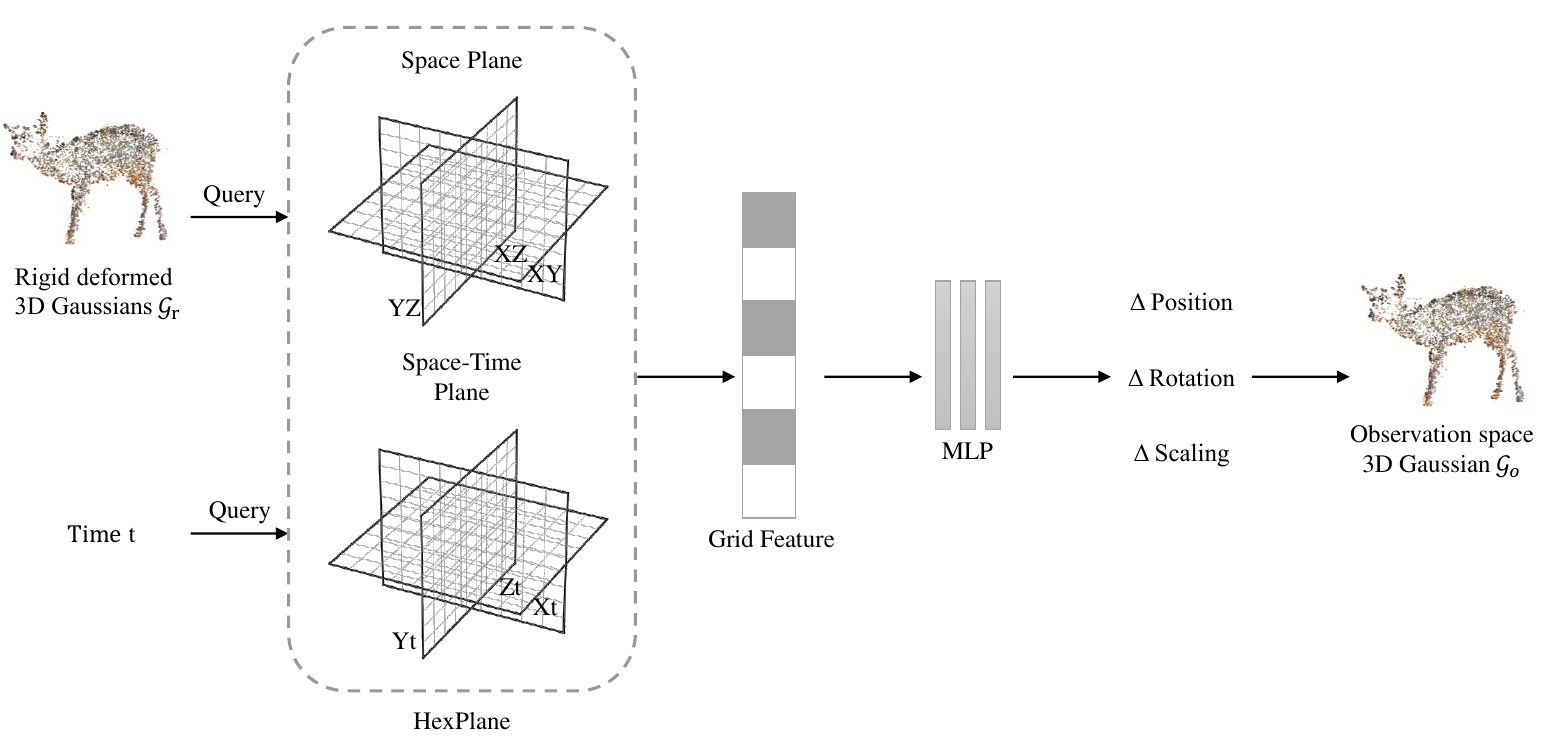}
    \caption{Illustration of the non-rigid deformation field using HexPlane, which captures intricate motion details.}
    \label{fig:hexplane}
\end{figure*}

\section{HexPlane Deformation Field} 
\label{sec:hexplane_deformation_field}

To achieve a precise refinement of the rigidly deformed 3D Gaussian \( \mathcal{G}_r \) into the observed 3D Gaussian \( \mathcal{G}_o \), we adopt a HexPlane combined with an MLP as a 3D Gaussian deformation model. This setup estimates the positional offset, rotational variation, and scaling adjustment of each Gaussian based on its spatial coordinates \((x, y, z)\) and temporal input \(t\). As depicted in \Cref{fig:hexplane}, the HexPlane framework breaks down the 4D field into six feature planes, each corresponding to a pair of coordinate axes. This decomposition method not only ensures computational efficiency but also represents the 4D field as a weighted combination of trainable 4D basis functions. We first extract feature representations from the HexPlane. These features are then processed by an MLP decoder, which outputs the Gaussian's positional displacement, rotation adjustment, and scaling transformation.  

\section{Implementation Details}
\label{sec:implementation_details}

We initialize the static object with 10000 Gaussian points within a sphere of radius 2 and train the object over 1500 steps. Using the Coverage Axis++ method \cite{wang2024coverage}, 70 skeleton points are extracted from the resulting static 3D Gaussian. Subsequently, skeleton poses are trained for 2,500 steps, with a smoothing window of size 3 applied to ensure temporal smoothness. The learning rate for pose training gradually decreases from 0.00005 to 0.000005. The learning rate for the deformation hexplane is initialized at $1.6 \times 10^{-4}$, and it decays to $1.6 \times 10^{-6}$ by the end of the training process. The loss functions are configured as follows: the weight for the SDS loss is fixed at 1, while the reconstruction and mask losses are weighted at $2 \times 10^4$ and $1 \times 10^3$, respectively. Real-time rendering achieves a performance rate of 150 FPS.

\section{Additional Information on Loss Functions}
\label{sec:loss_functions}

We adopt the multi-view Score Distillation Sampling (SDS) loss formulation as described in \cite{zeng2024stag4d}. At each timestep \( t \), we obtain six anchor views \( \{I_{t}^{i}\}_{i\in\{1...6\}} \) along with a reference view \( I_{t}^{ref} \). During optimization, we employ multi-view SDS, leveraging both the generated images \( \{I_{t}^{i}\}_{i = 1...6} \) and the reference image \( I_{t}^{ref} \). The multi-view score distillation loss function \( \mathcal{L}_{MV\text{-}SDS} \) is defined as:

\begin{align*}
\mathcal{L}_{MV\text{-}SDS} &= \alpha_{1}\mathcal{L}_{SDS}^{i} + \alpha_{2}\mathcal{L}_{SDS}^{ref} \\
&= \alpha_{1}\mathcal{L}_{SDS}(\phi, I_{t}^{i}) + \alpha_{2}\mathcal{L}_{SDS}(\phi, I_{t}^{ref}),
\end{align*}
where \( \alpha_{1} \) and \( \alpha_{2} \) are weighting parameters. The index \( i \) is chosen based on the proximity of the rendering viewpoint to that of the generated images. This selection process, known as multi-view score distillation sampling, involves identifying the reference image that is closest to the rendered camera view for SDS loss computation.

The gradient of the SDS loss is given by:

\[
\nabla_{\theta}\mathcal{L}_{SDS}(\phi,\mathbf{x}) = \mathbb{E}_{t,\epsilon}\left[\omega(t)(\hat{\epsilon}_{\theta}(\mathbf{z}_t;\mathbf{I}_{in},\mathbf{R},\mathbf{T},t)-\epsilon)\frac{\partial\mathbf{x}}{\partial\theta}\right],
\]

\[
\hat{z} = z_t - \sigma_t\hat{\epsilon}_t(z;\mathbf{I}_{in},\mathbf{R},\mathbf{T}).
\]
Here, \( \theta \) represents the parameters of the 3D representation, \( \mathbf{x} \) denotes the rendered image at the current view, \( t \) is the timestep in the diffusion process, \( \epsilon \) is the ground-truth noise, and \( \hat{\epsilon} \) is the predicted noise from the noisy image \( \mathbf{z}_t \), conditioned on the initial input \( \mathbf{I}_{in} \) and the relative camera pose \( (\mathbf{R},\mathbf{T}) \).

In addition, we compute both the reconstruction loss \( \mathcal{L}_{rec} \) and the foreground mask loss \( \mathcal{L}_{mask} \) using the reference image. These losses are formulated as follows:

\begin{equation}
    \mathcal{L}_{rec} = \left\| I_{t}^{i} - I_{t}^{ref} \right\|^2,
\end{equation}

\begin{equation}
    \mathcal{L}_{mask} = \left\| M_{t}^{i} - M_{t}^{ref} \right\|^2,
\end{equation}
where \( I_{t}^{i} \) and \( I_{t}^{ref} \) represent the generated and reference images, respectively, while \( M_{t}^{i} \) and \( M_{t}^{ref} \) denote their corresponding foreground masks.

To ensure spatiotemporal consistency, we apply Total Variation (TV) regularization \( \mathcal{L}_{reg} \), following \cite{cao2023hexplane}.

The final optimization objective is formulated as:

\begin{equation}
\mathcal{L} = \mathcal{L}_{MV-SDS} + \lambda_1 \mathcal{L}_{rec} + \lambda_2 \mathcal{L}_{mask} + \lambda_3 \mathcal{L}_{reg},
\end{equation}

where \( \lambda_{1} \), \( \lambda_{2} \), and \( \lambda_{3} \) are weighting parameters that control the contributions of the respective loss terms.

\section{Additional Results for 4D Generation}
\label{sec:additional_results}

\begin{figure*}[h]
    \centering
    \includegraphics[width=\linewidth]{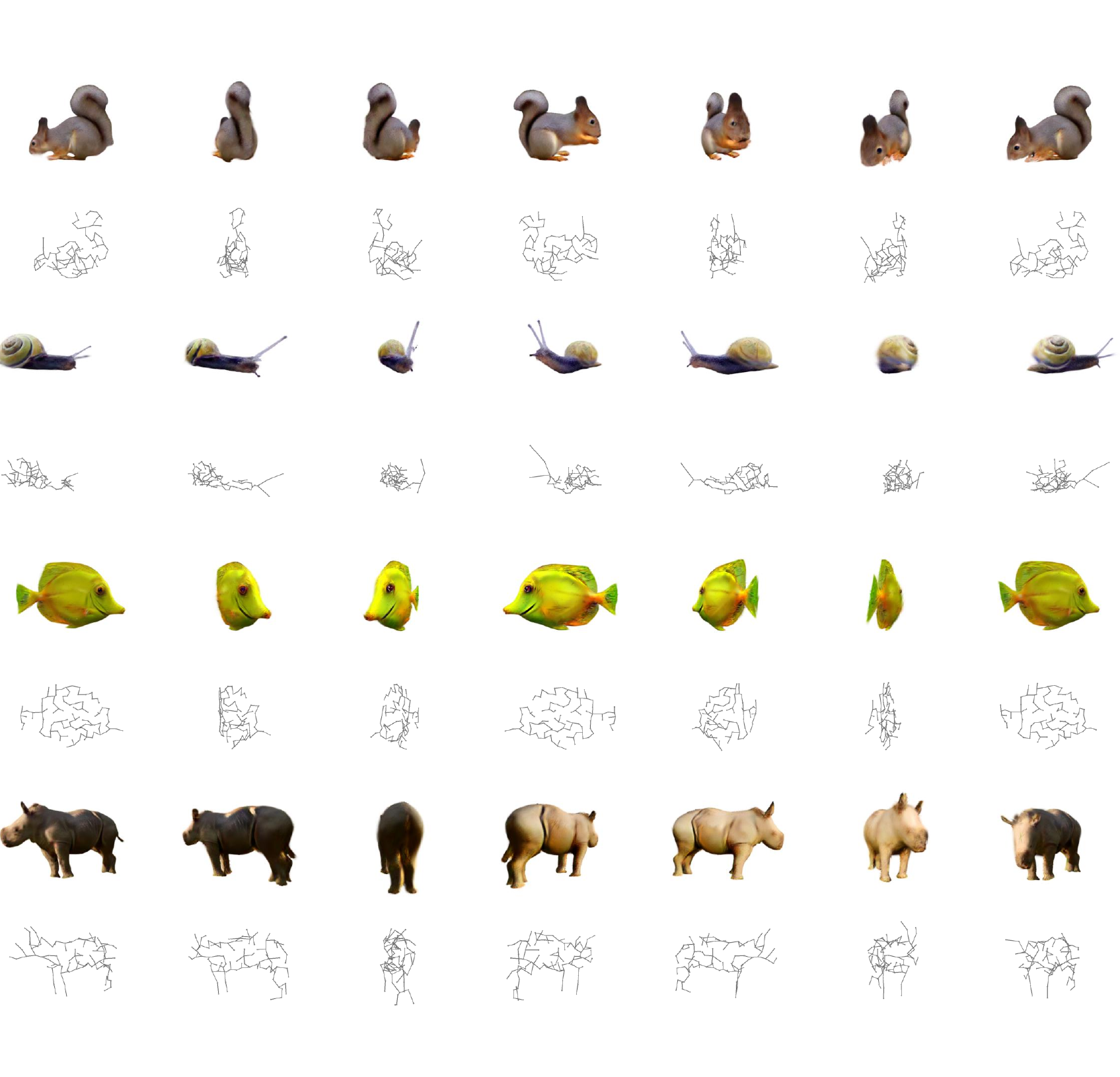}
    \caption{Qualitative results illustrating gradual changes in time stamps and view angles from left to right.}
    \label{fig:more_visualize_results_rotation_view}
\end{figure*}

\begin{figure*}[h]
    \centering
    \includegraphics[width=\linewidth]{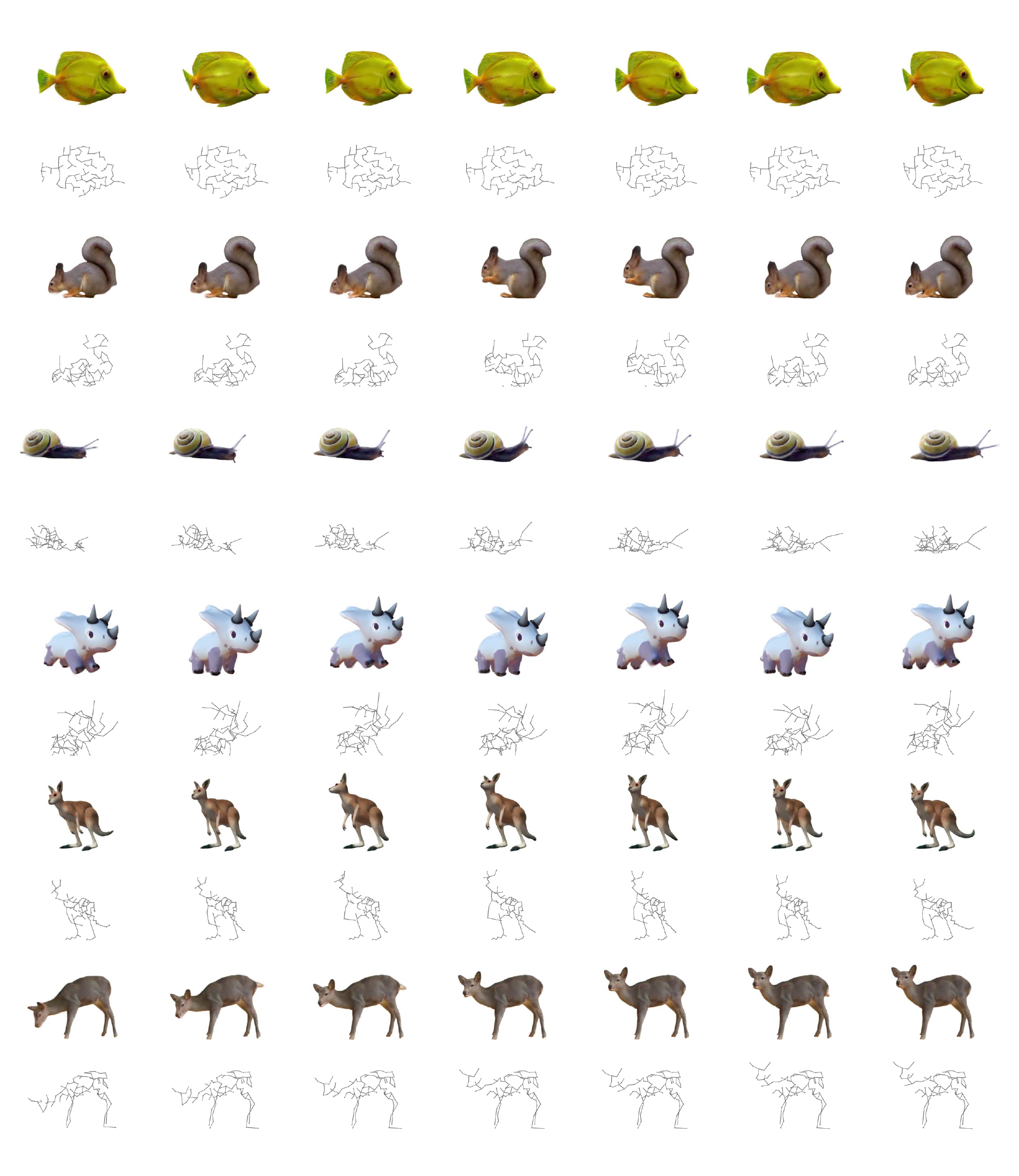}
    \caption{Front-view qualitative results with varying time stamps from left to right.}
    \label{fig:more_visualize_results_front_view}
\end{figure*}

We present additional results for 4D generation to further illustrate the effectiveness of our approach. Rotation-view visualizations of the generated objects are shown in \Cref{fig:more_visualize_results_rotation_view}, while front-view visualizations are provided in \Cref{fig:more_visualize_results_front_view}. Additionally, we include visualizations of the corresponding skeletons that capture the objects' motion. Notably, the skeleton poses align seamlessly with the objects' movements, highlighting the motion modeling ability of SkeletonGaussian.

\section{Failure Cases}
\label{sec:discussion_on_failed_cases}

\begin{figure*}[hb]
    \centering
    \includegraphics[width=\linewidth]{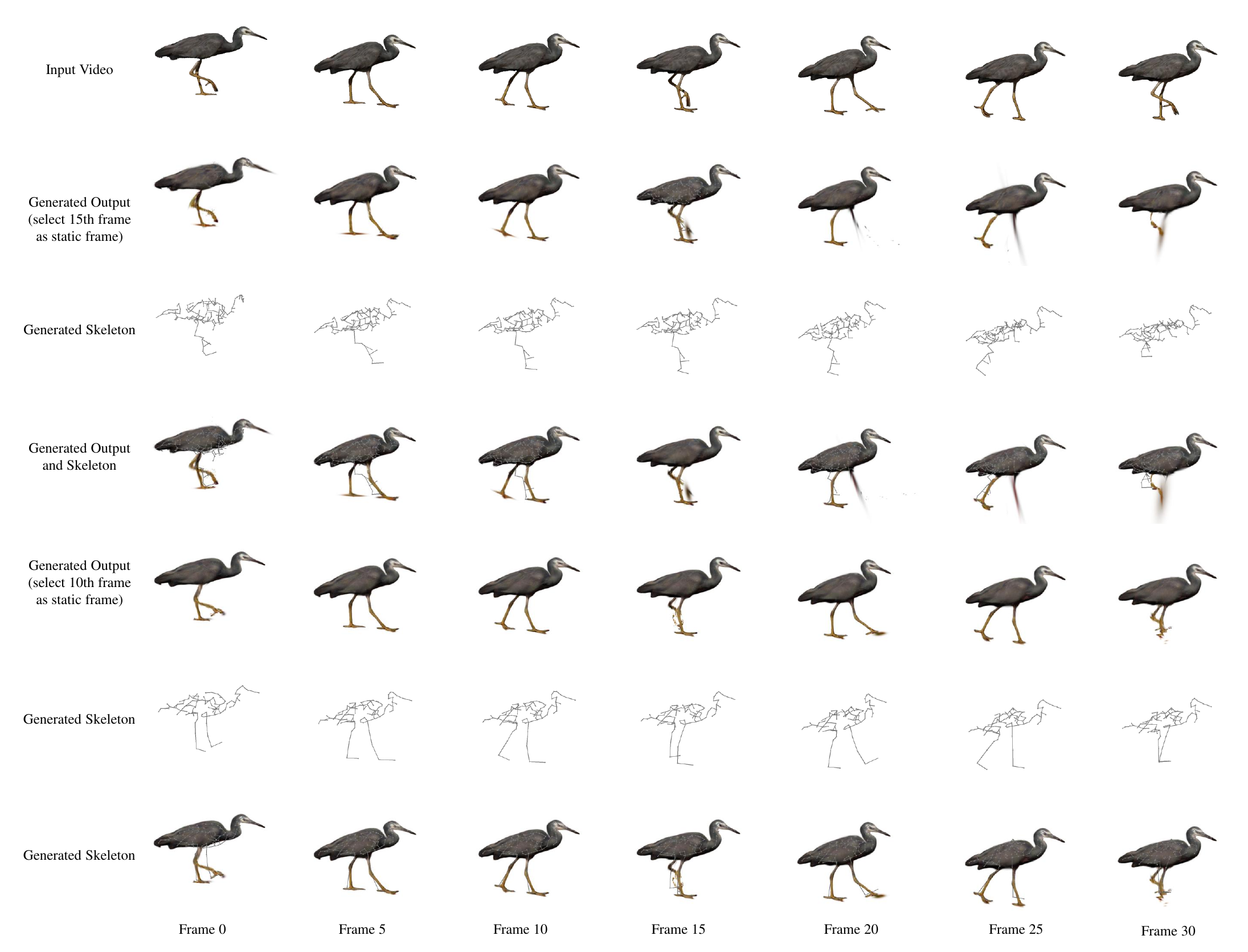}
    \caption{Visualization of failure cases. \textbf{Top:} The egret case shows how leg posture misestimation can lead to incorrect results when the 15th frame is selected as the static frame. \textbf{Bottom:} Selecting the 10th frame as static frame can resolves the issue.}
    \label{fig:failed_cases}
\end{figure*}

\begin{figure*}[hb]
    \centering
    \includegraphics[width=\linewidth]{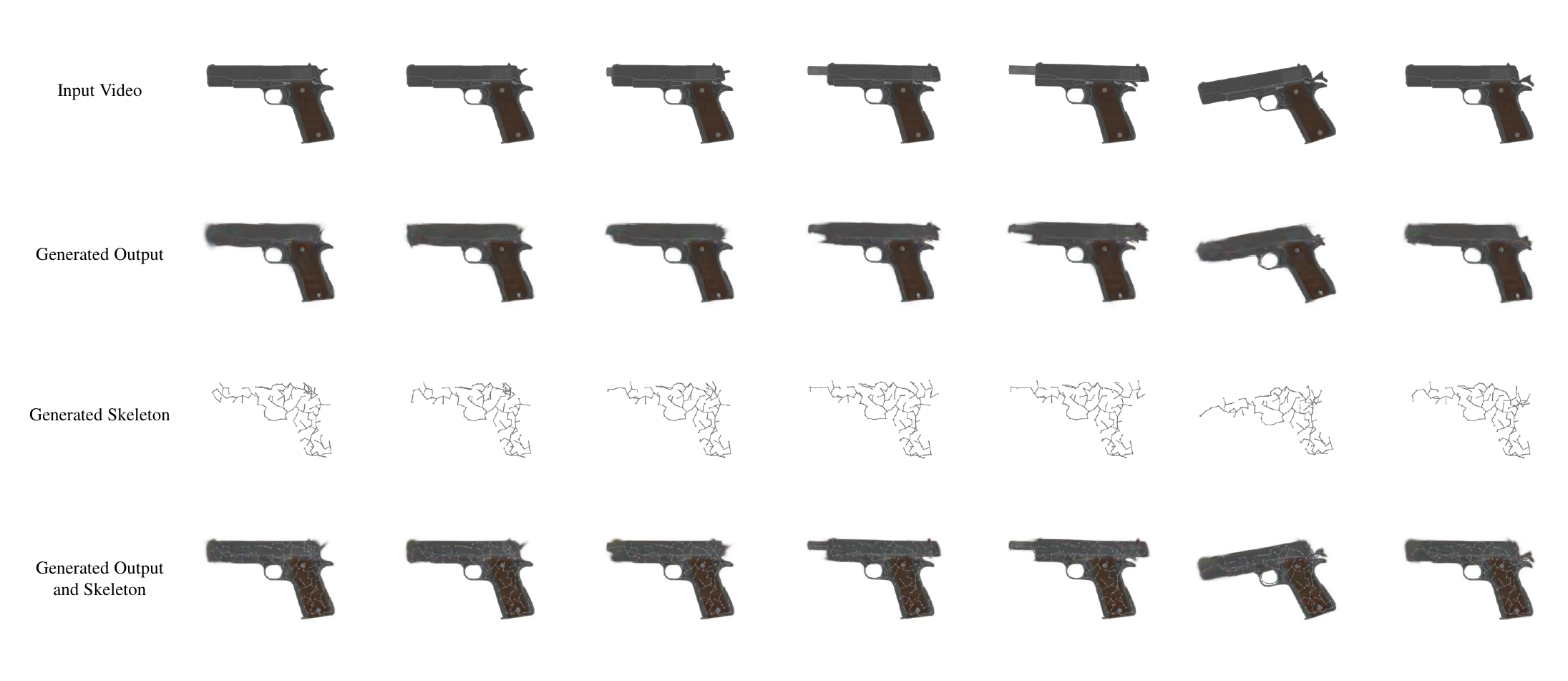}
    \caption{Visualization of failure cases. The pistol case demonstrates the challenges of modeling non-articulated structures.}
    \label{fig:failed_cases2}
\end{figure*}

Through experimental analysis, we observe that the quality of skeleton extraction significantly impacts our 4D generation performance, although these failure cases occur infrequently in our test scenarios. Our investigation identifies two primary categories of failures:

\textbf{Category 1: Inaccurate Skeleton Extraction} As shown in \Cref{fig:failed_cases}, the egret example illustrates how errors in skeleton extraction can degrade generation quality. In this case, the canonical frame is set to the 15th frame of the sequence, where the bird's legs are crossed. Consequently, our method misinterprets the leg topology, leading to incorrect skeletal connections that compromise the quality of 4D generation. This issue can be mitigated by selecting a different canonical frame (the 10th frame in this example) or manually refining the extracted skeleton.

\textbf{Category 2: Non-Skeletal Structures} The pistol example in \Cref{fig:failed_cases2} highlights the limitations of skeletal models in representing rigid-body transformations. Since the gun barrel lacks articulation, it does not conform to a skeletal parameterization, making this representation unsuitable for capturing its motion. As a result, our framework struggles to accurately reconstruct sliding motions along the barrel axis, as skeletal transformations cannot effectively approximate rigid translations. This limitation underscores the broader challenge of applying skeletal-based approaches to non-articulated objects. However, our method excels at modeling naturally articulated structures such as humans, animals, and plants, where motion priors align well with the skeletal representation space.